\documentclass[11pt]{article}

\usepackage[preprint]{acl}

\usepackage[utf8]{inputenc}
\usepackage[T1]{fontenc}
\usepackage{times}

\usepackage{latexsym}
\usepackage{graphicx}
\usepackage{booktabs}
\usepackage{array}
\usepackage{amsmath,amssymb}
\usepackage{xcolor}
\usepackage{url}
\title{Benefit and Cost of Random-Token Replacement under Repeated Data:\\
A Known-Entropy Decomposition and Evidence from Ten Real Corpora}

\author{Wonsup Shin \\
  Flitto \\
  \texttt{wonsup.shin@flitto.com} \\\And
  Jingu Kim \\
  Flitto \\
  \texttt{jin@flitto.com}}
\date{}

\begin{document}
\maketitle

\begin{abstract}
\noindent
When unique pretraining data is scarce, models see the same tokens many times and are thought to memorize. Random-token replacement, an input-corruption augmentation, is one remedy reported to lower validation loss, but a lower loss alone does not reveal which source of error it reduces. On a controlled source whose entropy rate is known in closed form, we show that replacement shrinks predictions toward the corpus average. We also introduce a decomposition that separates validation loss into approximation, estimation, and memorization terms. This shrinkage view predicts that replacement's bias cost needs no memorization but its benefit does. Memorization raises validation loss after its minimum, so we report the effect at two points, the minimum validation loss and the end of a fixed budget. Where replacement helps, it reduces this rise by about $60\%$. At the end of the budget, it turns from harmful to helpful as the controlled source becomes less predictable; at the minimum, it helps at no predictability level under the fixed recipe. On Korean and Japanese web text, replacement removes about a third of the loss rise after the minimum by the end of the budget, but its effect at the minimum is much smaller. Across ten real corpora, replacement lowers the final loss on nine. It harms only the census corpus, program-generated JSON that is the most redundant of the ten and much more redundant than any of 64 further prose and code corpora we measure.
\end{abstract}

\section{Introduction}
\label{sec:intro}

Pretraining a language model when the supply of unique text is limited~\citep{villalobos2022data} forces repeated passes over the same tokens. Beyond roughly four epochs the marginal value of repetition falls sharply, and past ${\sim}16$ epochs additional repetition adds essentially nothing~\citep{muennighoff2023scaling}. This is usually attributed to memorization, as the model overfits the training data.

A widely used remedy is training-time input-corruption augmentation. At each step it replaces a fraction of input tokens with random tokens or masks them~\citep{chen2026demystifying,xu2026dataconstrained}. Yet existing studies leave three questions. First, validation loss alone cannot tell memorization from other sources of error, so an improvement in it does not show whether the corruption acts on memorization or on something else. Second, it is unclear whether the cost and benefit of the augmentation appear at the same point of training, and so whether the effect depends on when training stops. Third, the evidence comes almost entirely from English prose and code~\citep{chen2026demystifying,xu2026dataconstrained}, leaving unclear how far it carries to other languages and other kinds of text. We address all three for random-token replacement.

\begin{figure*}[t!]
\centering
\includegraphics[width=0.92\textwidth]{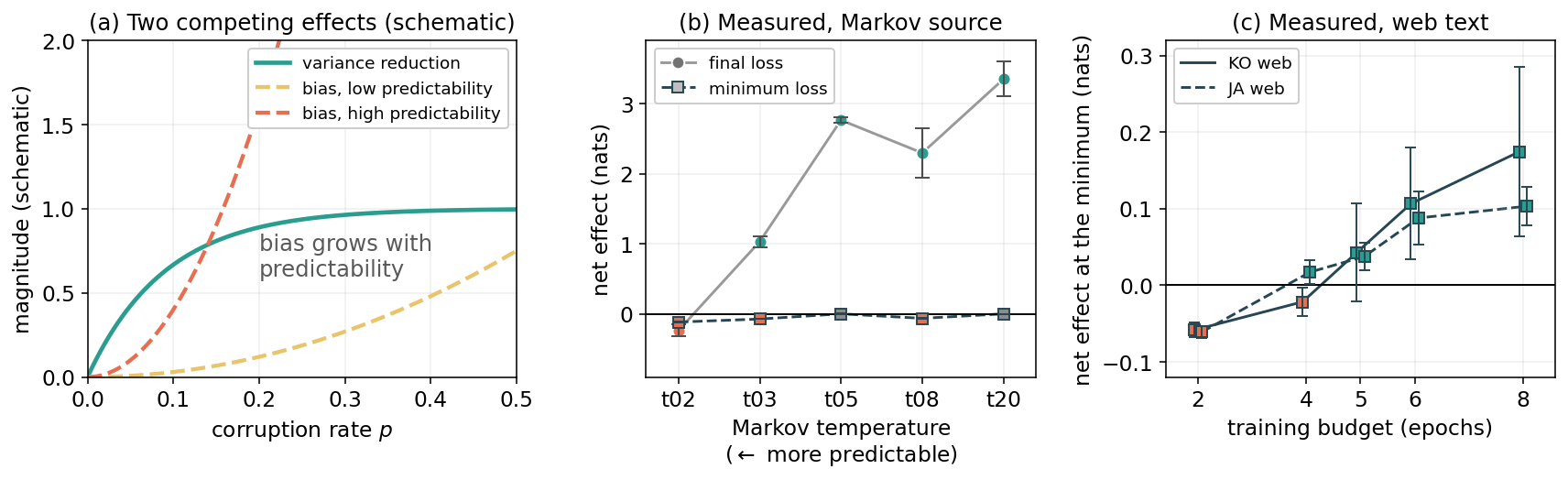}
\caption{(a) A schematic of the benefit and the bias as the corruption rate $p$ rises; the bias climbs faster on a more predictable corpus. (b, c) Measured net effect of replacement, positive when it helps. (b) The Markov source at temperatures $0.2$ to $2.0$, written $t02$ to $t20$, under a constant learning rate. (c) Web text under a learning rate decayed within each budget (Table~\ref{tab:budget}). Error bars give the s.d. over three seeds.}
\label{fig:mechanism}
\end{figure*}

Replacement shrinks the model's predictions toward the corpus average. Shrinkage reduces variance at the cost of a bias~\citep{bishop1995training,xie2017data}. Under repetition, the natural reading is that this variance comes from memorization. Where text is predictable, the next token is nearly determined, so the shrinkage should cost more and help less (Figure~\ref{fig:mechanism}(a)). From this shrinkage view we predict that the cost appears immediately, while the benefit emerges only after the model has memorized finite data, and that the cost is larger on more predictable text.

We derive this view on an order-1 Markov source, whose entropy rate is known in closed form (\S\ref{sec:mechanism}). On this source we introduce a decomposition of validation loss into approximation, estimation, and memorization terms. It uses the entropy rate as the ground-truth lower bound on loss, and it identifies which term produces the benefit or harm (\S\ref{sec:mem}).

We then test the range of the effect on ten real corpora while holding the model and training recipe fixed.\footnote{Code and configuration files are available at \url{https://github.com/wonsupshin-flitto/random-token-replacement-repeated-data}.} The corpora span four respects: language, modality, structure, and origin (\S\ref{sec:setup}). We repeat the core comparison at two model scales.

On our main training budget (\S\ref{sec:setup}), we report each effect at two points of training, since the two can disagree. The \emph{minimum loss} is the lowest validation loss a run reaches, the point early stopping selects, and the \emph{final loss} is the validation loss at the end of a fixed budget. Our main budget runs far past the minimum as a deliberate stress test. Runs are often not stopped at the minimum loss, since they follow a fixed token budget~\citep{hoffmann2022chinchilla,muennighoff2023scaling} and continued training can improve downstream performance after the pretraining loss stops falling~\citep{liu2023same}.

\paragraph{Contributions.}\mbox{}

\textbf{Our decomposition of validation loss shows what an augmentation removes.} On a source of known entropy rate, we derive the shrinkage view of replacement (\S\ref{sec:mechanism}) and decompose validation loss into approximation, estimation, and memorization terms (\S\ref{sec:mem}). The decomposition needs only validation losses on such a source, and nothing in it is specific to replacement, so it applies to any augmentation.

\textbf{Replacement helps by reducing the loss rise after the minimum, so its value depends on when training stops.} Where replacement helps, it reduces the memorization term by ${\sim}60\%$ while the other two terms offset each other (\S\ref{sec:mem}). Its bias cost is present even when nothing is memorized, but the benefit follows only after memorization accumulates. The effect on the controlled source therefore turns from harm to benefit on final loss as predictability falls, while replacement lowers the minimum loss at no predictability level under the fixed recipe (Figure~\ref{fig:mechanism}(b)). On Korean and Japanese web text, replacement removes a third or more of the overfitting gap, with or without learning-rate decay, but its effect at the minimum is much smaller (Figure~\ref{fig:mechanism}(c), \S\ref{sec:safe}).

\textbf{Replacement harms only an extremely redundant corpus in our tests.} Beyond the ten corpora we train on, we measure a compression-based redundancy, the gzip ratio, on 64 corpora that are not program-generated. On final loss, replacement helps every trained corpus that is not program-generated and harms only the census corpus, census records serialized to JSON. Mutual information does not set this corpus apart from the helped ones, but by gzip it is more redundant than all 64 surveyed corpora (\S\ref{sec:redundancy}).
\section{Related Work}
\label{sec:related}

\paragraph{Training-time augmentation.}
Corruption has a long history as a pretraining objective~\citep{devlin2019bert,lewis2020bart}. We use it differently. The target stays the clean next token and the corruption acts on the input alone. Recent data-constrained studies use masking, replacement and infilling this way, as regularizers under repeated causal-LM training~\citep{chen2026demystifying,xu2026dataconstrained}. \citet{chen2026demystifying} report that replacement is the strongest of the three on English.

\paragraph{Measuring what a model has learned.}
\citet{jumelet2023transparency} train on artificial language-like data from a probabilistic grammar, where the true distribution gives an exact lower bound on loss. We use the same access for a different end, splitting loss into terms rather than scoring how close a model gets to the bound. \citet{bahri2024explaining} also decompose loss, but to explain how it scales with model and data size, whereas ours splits loss at one model size. Our memorization term, a rise in held-out loss after a run's minimum, is narrower than the exact reproduction of training strings, which \citet{tirumala2022memorization} show can begin before a model overfits.

\paragraph{Noise as regularization.}
That input noise acts as a smoothing regularizer is classical. \citet{bishop1995training} shows it is equivalent to Tikhonov regularization, and \citet{xie2017data} give a closed-form view of token noising as interpolation toward lower-order distributions. Our mechanism (\S\ref{sec:mechanism}) is an instance of this view. Augmentation is also known to help in moderation and hurt when it erases structure~\citep{lin2024spectral,nguyen2025coarse}. We locate the point where the predicted bias starts to outweigh the benefit, on a source of known entropy rate.

\paragraph{Early stopping and fixed budgets.}
Studies of repeated data read loss at different points of training. \citet{kaplan2020scaling} stop early, while \citet{muennighoff2023scaling} deliberately do not, so that overfitting stays visible. The two conventions can disagree, since validation loss need not fall monotonically with training time~\citep{nakkiran2020double}. \citet{muennighoff2023scaling} report that loss rises again with more epochs, expect regularization to prevent that rise, and leave the case outside their scaling law. A decay schedule should also be matched to the training budget~\citep{hoffmann2022chinchilla}, so we test that expectation under a matched schedule as well as a constant rate, and report both points for every experiment on the main budget.

\paragraph{Non-English, speech, and tabular corpora.}
\citet{seto2024bilingual} study data-constrained pretraining in a targeted non-English language, and using ASR transcripts for pretraining has precedent~\citep{pelloin2022asr,sundararaman2021phonemebert}, where spoken and written text differ systematically in register~\citep{biber1988variation}. Transformers have also been trained on tabular data with one embedding per field~\citep{padhi2021tabular}, whereas we serialize tables to plain JSON text.

\paragraph{Predicting behavior from corpus statistics.}
The gzip ratio predicts data-dependent scaling~\citep{pandey2024gzip}, redundancy predicts memorization~\citep{zhang2025redundancy}, and \citet{jiang2026beyond} predict knowledge retention before training from a size-dependent mutual information score. Language modeling and compression are near-equivalent~\citep{deletang2024compression}, so such a scalar partly restates model loss. We ask instead whether such scalars predict when replacement helps (\S\ref{sec:redundancy}).

\section{Experimental Setup}
\label{sec:setup}

\paragraph{Corpora.}
We use ten real corpora and one controlled source. Each real corpus is labeled natural language or structured text. Independently, any corpus, real or controlled, is labeled program-generated when a schema, a template, or a random process produced it.

The natural-language corpora are Korean and Japanese web text from FineWeb-2~\citep{penedo2025fineweb2} and Korean speech transcripts from the open YODAS corpus~\citep{li2023yodas}. Each of the three has exactly $7.4$M training tokens. The structured corpora are source code from The Stack~\citep{kocetkov2022stack}, chess move sequences~\citep{chessgames}, SQL from SQL-create-context~\citep{sqlcreatecontext} in both a combined context-plus-query form (SQL-comb) and a context-only form (SQL-ctx), a system-log corpus from LogHub~\citep{zhu2023loghub}, and two tables serialized to JSON. One is a census dataset~\citep{kohavi1996adult}, highly templated and repetitive by construction, and the other the public road-safety accident table~\citep{grinsztajn2022tabular}. Code, the system-log corpus and the road-safety corpus also hold $7.4$M tokens; chess, the two SQL forms and the census corpus range from $2.3$M to $5.5$M, since most of these public datasets are small.

The controlled source is an order-1 Markov source whose true entropy rate $H$ is known in closed form from the transition matrix~\citep{cover2006elements}, sampled at temperatures from $0.2$ to $2.0$; the temperature sets $H$. Each temperature is a separate source. The controlled runs use $2$M training tokens and $8{,}000$ steps.

The program-generated label covers the census corpus, the two SQL forms, the Markov source, and the road-safety corpus and system-log corpus. We use the labels to describe corpora, not to explain the sign of the effect.

\begin{table*}[t]
\centering
\small
\begin{tabular}{lccccc}
\toprule
Corpus & Gain (\%) & $\Delta L_{\text{final}}$ & $L_{\text{min}}$ & $\Delta L_{\text{min}}$ & epoch at min \\
\midrule
KO web & $35.2{\pm}1.5$ & $+0.93{\pm}0.12^{*}$ & $4.91\to4.85$ & $+0.06{\pm}0.08$ & $4.4\to5.3$ \\
JA web & $37.5{\pm}1.7$ & $+1.14{\pm}0.05^{*}$ & $4.70\to4.61$ & $+0.09{\pm}0.01^{*}$ & $4.4\to5.5$ \\
KO transcript & $40.1{\pm}0.7$ & $+0.93{\pm}0.01^{*}$ & $3.37\to3.31$ & $+0.06{\pm}0.00^{*}$ & $4.0\to5.8$ \\
code & $26.3{\pm}1.4$ & $+0.24{\pm}0.04^{*}$ & $1.86\to1.92$ & $-0.06{\pm}0.03$ & $4.4\to5.8$ \\
chess & $25.9{\pm}4.2$ & $+0.29{\pm}0.04^{*}$ & $0.84\to0.81$ & $+0.03{\pm}0.01^{*}$ & $3.8\to7.6$ \\
road-safety (JSON)$^{\dagger}$ & $41.9{\pm}1.3$ & $+0.04{\pm}0.00^{*}$ & $0.17\to0.16$ & $+0.00{\pm}0.00$ & $6.0\to7.7$ \\
\midrule
SQL-comb & $-4.4{\pm}9.4$ & $+0.04{\pm}0.04$ & $1.44\to1.38$ & $+0.06{\pm}0.05$ & $6.6\to7.3$ \\
SQL-ctx & $-15.2{\pm}16.4$ & $+0.05{\pm}0.09$ & $2.40\to2.27$ & $+0.13{\pm}0.02^{*}$ & $7.1\to9.3$ \\
census (JSON) & $2.0{\pm}22.9$ & $-0.06{\pm}0.02^{*}$ & $0.26\to0.32$ & $-0.06{\pm}0.02$ & $2.1\to1.2$ \\
loghub$^{\dagger}$ & $13.7{\pm}7.2$ & $+0.07{\pm}0.01$ & $0.93\to0.91$ & $+0.02{\pm}0.02$ & $6.3\to4.3$ \\
\bottomrule
\end{tabular}
\caption{Effect of replacement at 336M, mean${\pm}$s.d. over three seeds. $\Delta$ is $L_{\textsc{none}}-L_{\textsc{replace}}$ in nats, positive when replacement helps; $^{*}$ is $p<0.05$ in a one-sample $t$-test. Gain (\%) is the regularization gain of \S\ref{sec:setup}, and $L_{\text{min}}$ and its epoch read \textsc{none}$\to$\textsc{replace}. Below the line, Gain does not differ from zero at $p<0.05$. $^{\dagger}$Two seeds.}
\label{tab:seed}
\end{table*}

\paragraph{Model and recipe.}
All runs train a decoder-only Transformer from scratch for $10{,}000$ steps, our main training budget, under one fixed recipe with a constant learning rate (App.~\ref{app:repro}). That is roughly $22$ epochs on each $7.4$M-token corpus, well past the point where repetition stops helping (App.~\ref{app:wall}). Every baseline reaches its minimum loss within its first eight epochs (Table~\ref{tab:seed}), so each corpus, large or small, spends most of the budget past that point. The main model has 336M parameters and the scale check a 134M variant.

\paragraph{Input-corruption augmentation.}
We study random-token replacement, which independently replaces each input token with a uniformly random vocabulary token at rate $p$, leaving the prediction targets intact. Unless noted, all main experiments use $p{=}0.15$; we write \textsc{replace} for this condition and \textsc{none} for the no-augmentation baseline with $p{=}0$. The corpus marginal, the average token distribution, does not enter the noise. It enters only when the model averages over the clean token that a corrupted context could have come from (\S\ref{sec:mechanism}).

\paragraph{Metrics.}
We report validation loss in nats throughout. Each corpus reserves a disjoint held-out validation slice, on which every loss is measured (App.~\ref{app:repro}). We evaluate every 300 steps, or every 250 on the Markov source, and read two points from each run. The minimum loss $L_{\text{min}}$ is the lowest validation loss of the run, reported with the epoch at which it occurs. Early stopping with a patience of five evaluations selects this same point on 172 of the 174 runs whose minimum we report (App.~\ref{app:es}). The final loss $L_{\text{final}}$ is the last validation loss of the run. At either point, the effect of replacement is $\Delta = L_{\textsc{none}} - L_{\textsc{replace}}$, so a positive sign means replacement helps. The \emph{overfitting gap} of a run is $g = L_{\text{final}} - L_{\text{min}}$, and the \emph{regularization gain} of replacement is $(g_{\textsc{none}} - g_{\textsc{replace}})/g_{\textsc{none}}$, the fraction of the overfitting gap that replacement removes. As redundancy measures we report the gzip compression ratio of the raw text, lower for more redundant text, and the bigram mutual information (MI) normalized by unigram entropy. Both vary with the writing system, so we compare corpora on them only within one writing system.

\section{Shrinkage and Its Bias Cost}
\label{sec:mechanism}

On an order-1 source, each token depends only on the token before it, its context, and replacement swaps the context for a random token with probability $p$. The noise is uniform over the vocabulary, yet the model is pulled toward the corpus marginal. A model that sees a possibly corrupted context $\tilde c$ does best by averaging over the clean context that could have produced it,
\begin{align}
Q_{\tilde c} &= \alpha_{\tilde c}\, P_{\tilde c} + (1-\alpha_{\tilde c})\, \pi \label{eq:shrink}\\
\alpha_{\tilde c} &= \frac{(1-p)\,\pi(\tilde c)}{(1-p)\,\pi(\tilde c) + p\,q(\tilde c)} \label{eq:alpha}
\end{align}
where $Q_{\tilde c}$ is the best prediction after seeing $\tilde c$, $\alpha_{\tilde c}$ the probability that $\tilde c$ is clean, $P_{\tilde c}$ the true conditional distribution, $\pi$ the corpus marginal, and $q$ the noise distribution. The marginal appears in Eq.~\eqref{eq:shrink} for any $q$, because it comes from averaging the clean context under $\pi$. The noise $q$ only sets $\alpha_{\tilde c}$ in Eq.~\eqref{eq:alpha}. This averaging is the shrinkage view of input noising~\citep{bishop1995training,xie2017data}. Shrinkage reduces variance at the cost of a bias. Under repetition we expect the variance it removes to come from memorization. By Eq.~\eqref{eq:shrink}, the prediction misses the true conditional by $Q_{\tilde c}-P_{\tilde c}=(1-\alpha_{\tilde c})(\pi-P_{\tilde c})$. This gap is the bias. It grows with how far the true conditional $P_{\tilde c}$ departs from the marginal $\pi$, that is, with how predictable the next token is from its context, which the size of $\pi-P_{\tilde c}$ measures. The other factor, $1-\alpha_{\tilde c}$, sets how often the shrinkage lands on a clean context. It depends on whether the noise draws a token that occurs in the corpus. On the Markov source it almost always does, so shrinkage falls on every clean context. On the real corpora it often does not, so the corrupted position is recognizable and much less shrinkage falls on clean contexts (App.~\ref{app:alpha}). Turning to the benefit, it also varies with the source. The loss rise from memorization should grow with the entropy rate $H$, so on a highly predictable source replacement removes little. As predictability rises, the benefit shrinks and the bias grows. The sign of the effect depends on which of the two is larger.

The two also act at different times. The bias is present from the start, while the variance reduction can matter only after the minimum loss, once the model starts to overfit. At the minimum the bias is therefore exposed and the net effect is small or negative; by the end of the budget the variance reduction dominates on all but the most predictable sources (Figure~\ref{fig:mechanism}(c)).

\section{The Benefit of Replacement Lies in the Memorization Term}
\label{sec:mem}

\S\ref{sec:mechanism} expects the variance that replacement removes to come from memorization. A lower validation loss alone cannot test this, because it does not show whether replacement helps the model learn the source or reduces the loss rise that comes with memorizing the training sample. We separate the two on the Markov source, whose entropy rate is known.

Starting from the entropy rate $H$, the lower bound on loss~\citep{cover2006elements}, we write the final loss of a run as
\begin{equation}
\begin{aligned}
&L_{\text{final}} = H + \underbrace{(L_{\text{abund}}-H)}_{\text{APPROX}} \\
&\quad + \underbrace{(L_{\text{min}}-L_{\text{abund}})}_{\text{EST}} + \underbrace{(L_{\text{final}}-L_{\text{min}})}_{\text{MEM}}
\end{aligned}
\label{eq:decomp}
\end{equation}
where $L_{\text{abund}}$ is the loss that a model of the same capacity reaches with abundant data (App.~\ref{app:repro}), and $L_{\text{min}}$ is the minimum loss that the same model reaches on the small corpus of \S\ref{sec:setup}. These three operational terms split the excess $L_{\text{final}}-H$ at these two reference losses. The APPROX term is the loss above $H$ that the model keeps even with abundant data. The EST term is the further loss at the minimum because the data are limited. MEM equals the overfitting gap $g$ of \S\ref{sec:setup} measured on this source. We attribute MEM to memorization, and the training loss supports this reading. Every run drives its training loss far below $H$, the lowest loss any model can reach on held-out sequences from the source (App.~\ref{app:memprobe}). Replacement leaves the training loss about the same or lower and raises the share of 50-token training spans the model reproduces verbatim~\citep{carlini2023quantifying}, so it does not stop the model from memorizing but reduces the rise in held-out loss that accompanies it. The abundant-data run, otherwise identical, rises by at most $0.004$ nats after its minimum (App.~\ref{app:repro}), so the rise appears only when the data repeat. The bias of replacement needs no memorization, so it does not show in $g$. Since nothing is memorized in the abundant-data run, the rise in APPROX under replacement, $\mathrm{APPROX}_{\textsc{replace}}-\mathrm{APPROX}_{\textsc{none}}$, measures this bias. It is $0.02$ to $0.05$ nats across the three sources (App.~\ref{app:repro}), so without scarcity replacement only costs, though we do not rule out a smaller optimization effect. At the minimum of a data-constrained run, a change in EST adds to it.

\begin{figure}[tb]
\centering
\includegraphics[width=\linewidth]{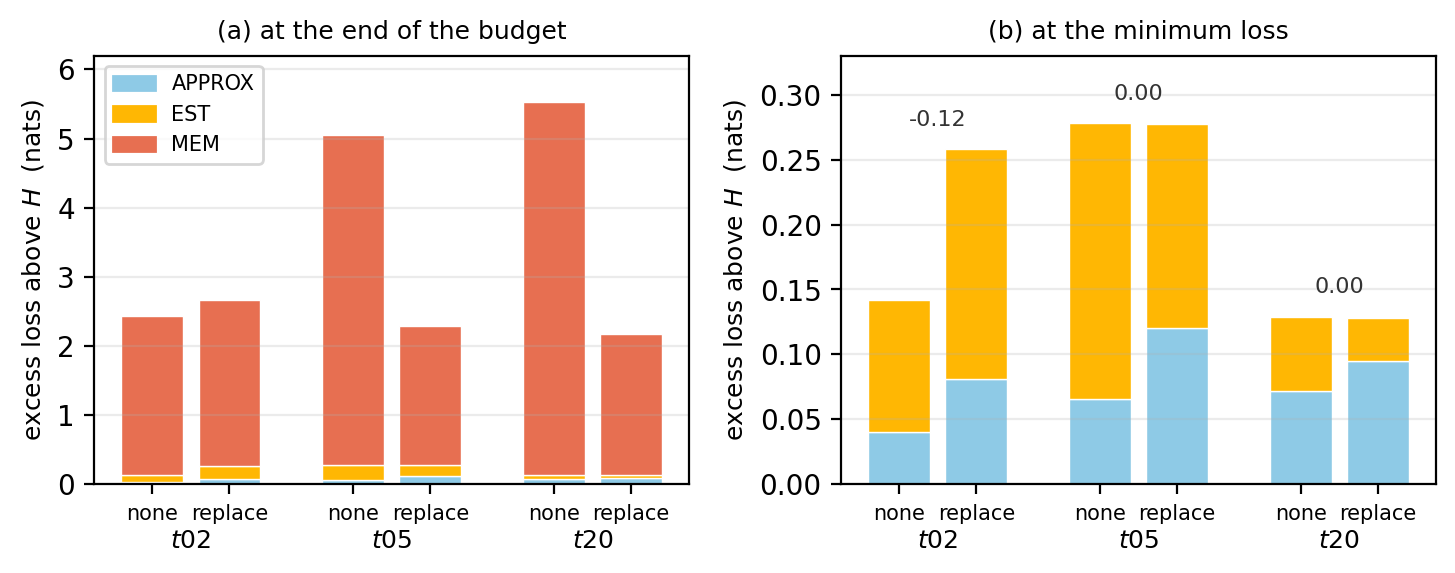}
\caption{Decomposition of Eq.~\eqref{eq:decomp} on the Markov source. The number above each pair in (b) is $\Delta L_{\text{min}}$.}
\label{fig:memdecomp}
\end{figure}

\paragraph{Replacement targets MEM.}
Figure~\ref{fig:memdecomp} shows the decomposition of Eq.~\eqref{eq:decomp} at three temperatures, $0.2$, $0.5$, and $2.0$, written $t02$, $t05$, and $t20$ (App.~\ref{app:decomp}). In the helpful regime, $t05$ and $t20$, replacement reduces MEM by $58$--$62\%$. At both sources the minimum loss is unchanged, with the rise in APPROX offset by a fall in EST, so the whole benefit lies in MEM (Figure~\ref{fig:memdecomp}(b)). At the most predictable source $t02$, by contrast, replacement does not reduce MEM and instead raises the minimum loss by $0.12$ nats. Both agree with \S\ref{sec:mechanism}, which predicts that on a highly predictable source replacement removes little of the loss rise from memorization and its bias is large. With no reduction in MEM to offset this bias, replacement harms.

\begin{figure*}[t]
\centering
\includegraphics[width=0.88\textwidth]{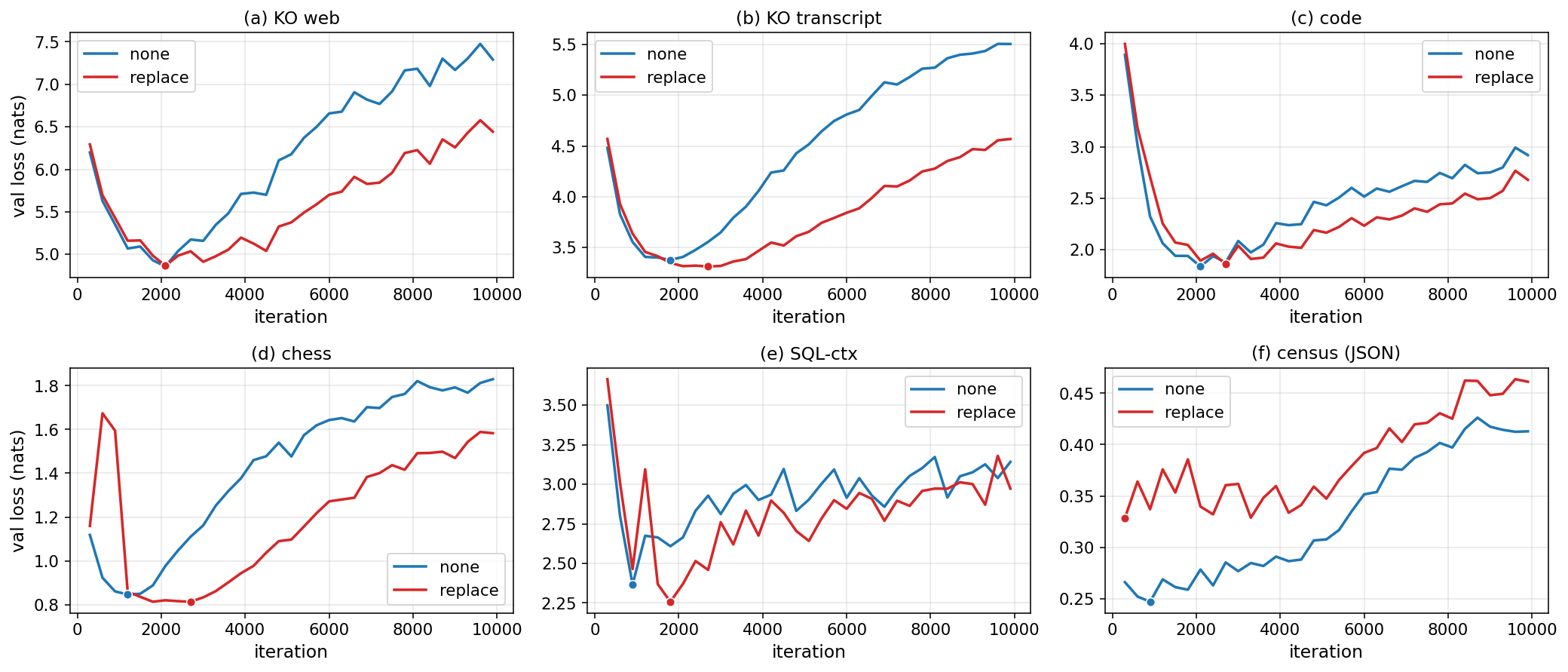}
\caption{Validation loss over training for one seed, \textsc{replace} in red and \textsc{none} in blue, with a dot at each run's minimum. Each panel has its own scale.}
\label{fig:curves}
\end{figure*}

\paragraph{The benefit scales with the entropy rate.}
Varying the Markov temperature sweeps the entropy rate from near-deterministic to near-uniform. The net effect of replacement on final loss, $L_{\textsc{none}}-L_{\textsc{replace}}$, increases overall along this axis (Figure~\ref{fig:mechanism}(b), with per-seed values in App.~\ref{app:seeds}). At the most predictable source replacement is harmful; as $H$ rises it becomes helpful and then strongly so, from $-0.23$ nats to $+3.35$ nats. A \emph{predictability threshold} therefore separates help from harm on this source, as the shrinkage view of \S\ref{sec:mechanism} predicts. At the minimum loss, by contrast, replacement helps at no temperature under the fixed recipe (Figure~\ref{fig:mechanism}(b)), as \S\ref{sec:mechanism} predicts for the point where the bias is exposed. A schedule sized to each of five budgets, from $4$ to $66$ epochs, improves the minimum by at most $0.023$ nats (App.~\ref{app:mkbudget}).

\section{Two Points of Training on Real Corpora}
\label{sec:safe}

On the Markov source, replacement helps on final loss but not at the minimum loss (\S\ref{sec:mem}). We ask whether the two points split the same way on real corpora.

Replacement lowers the final loss on web text, transcripts, code, chess, the system-log corpus, and the road-safety corpus (Table~\ref{tab:seed}, per-seed losses in App.~\ref{app:seeds}). On Korean and Japanese web and on Korean transcripts it removes about a third of the overfitting gap, with every seed positive and a one-sample $t$-test significant at $p<0.05$ for each corpus. On the two SQL forms the effect is within noise, and on the census corpus replacement is harmful, a case \S\ref{sec:redundancy} examines.

\paragraph{The minimum differs from the Markov source.}
Unlike on the Markov source, replacement often lowers the minimum loss a little on real corpora, significantly on Japanese web, transcripts, chess, and context-only SQL (Table~\ref{tab:seed}). We noted a reason for this difference in \S\ref{sec:mechanism}. On real corpora the noise often draws a token that never occurs in the corpus, so the average shrinkage weight on clean contexts $(1-\alpha_{\tilde c})$ is at most $0.06$, against $0.14$ to $0.15$ on the Markov source, close to its ceiling at the corruption rate $p{=}0.15$ (App.~\ref{app:alpha}). This smaller bias agrees with replacement lowering the minimum.

The effect at the minimum is still much smaller than on final loss. On web text, transcripts, code, and chess, the \textsc{none} and \textsc{replace} runs reach their minima at nearly the same loss and part only after it (Figure~\ref{fig:curves}(a--d)), and replacement delays the minimum by about one to four epochs (Table~\ref{tab:seed}), which suggests that the same training tokens stay useful for longer.

\begin{table}[t]
\centering
\small
\resizebox{\columnwidth}{!}{%
\begin{tabular}{llcccc}
\toprule
Corpus & Schedule & $g_{\textsc{none}}$ & $g_{\textsc{replace}}$ & Gain (\%) & $\Delta L_{\text{min}}$ \\
\midrule
KO web & constant & $2.46$ & $1.60$ & $35.1{\pm}0.5$ & $+0.07{\pm}0.06$ \\
KO web & cosine & $2.18$ & $1.29$ & $40.9{\pm}1.3$ & $+0.06{\pm}0.08$ \\
JA web & constant & $2.81$ & $1.76$ & $37.4{\pm}0.6$ & $+0.10{\pm}0.02$ \\
JA web & cosine & $2.62$ & $1.50$ & $42.9{\pm}1.1$ & $+0.10{\pm}0.04$ \\
\bottomrule
\end{tabular}}
\caption{Overfitting gap $g$ with and without replacement, and change in the minimum loss $\Delta L_{\text{min}}$, in nats under a constant and a cosine learning rate, 336M, three seeds.}
\label{tab:decay}
\end{table}

\paragraph{Learning-rate decay shrinks the gap only slightly.}
A constant learning rate might itself create the late rise that replacement removes. We repeat the web comparison with a cosine schedule that decays over the full budget to one tenth of the peak (Table~\ref{tab:decay}). Learning-rate decay shrinks the overfitting gap by only $7$ to $12\%$, so most of it remains, and replacement removes a larger share of the gap under learning-rate decay than under a constant rate. The effect at the minimum is the same under either schedule, within $0.01$ nats. Under stronger weight decay, an explicit regularizer, replacement still removes much of the gap (App.~\ref{app:wd}).

\begin{table}[t]
\centering
\small
\begin{tabular}{lcc}
\toprule
Budget & KO web & JA web \\
\midrule
$2$ epochs & $-0.06{\pm}0.01^{*}$ & $-0.06{\pm}0.01^{*}$ \\
$4$ epochs & $-0.02{\pm}0.02$ & $+0.02{\pm}0.02$ \\
$5$ epochs & $+0.04{\pm}0.06$ & $+0.04{\pm}0.02$ \\
$6$ epochs & $+0.11{\pm}0.07$ & $+0.09{\pm}0.03$ \\
$8$ epochs & $+0.17{\pm}0.11$ & $+0.10{\pm}0.03^{*}$ \\
best budget & $+0.16{\pm}0.10$ & $+0.08{\pm}0.01^{*}$ \\
\bottomrule
\end{tabular}
\caption{Change in the minimum loss, $\Delta L_{\text{min}}$ in nats, with the cosine schedule decayed within each budget, 336M, mean${\pm}$s.d. over three seeds; $^{*}$ is $p<0.05$. The best-budget row is $\min_b L^{\textsc{none}}_{\text{min}}(b) - \min_b L^{\textsc{replace}}_{\text{min}}(b)$ per seed, with $b$ over the five budgets.}
\label{tab:budget}
\end{table}

\begin{table}[tb]
\centering
\small
\resizebox{\columnwidth}{!}{%
\begin{tabular}{lcccc}
\toprule
 & \multicolumn{2}{c}{Gain (\%)} & \multicolumn{2}{c}{$\Delta L_{\text{min}}$} \\
\cmidrule(lr){2-3}\cmidrule(lr){4-5}
Corpus & 134M & 336M & 134M & 336M \\
\midrule
KO web & $52.8{\pm}1.4$ & $35.2{\pm}1.5$ & $+0.15{\pm}0.02^{*}$ & $+0.06{\pm}0.08$ \\
KO transcript & $61.5{\pm}2.1$ & $40.1{\pm}0.7$ & $+0.09{\pm}0.02^{*}$ & $+0.06{\pm}0.00^{*}$ \\
code & $44.4{\pm}2.7$ & $26.3{\pm}1.4$ & $+0.01{\pm}0.01$ & $-0.06{\pm}0.03$ \\
\bottomrule
\end{tabular}}
\caption{Regularization gain and change in the minimum loss, $\Delta L_{\text{min}}$ in nats, at two model scales, mean${\pm}$s.d. over three seeds; $^{*}$ is $p<0.05$.}
\label{tab:scale}
\end{table}

\paragraph{The best budget still favors replacement.}
A practitioner would rarely train this far past the minimum, and would instead choose fewer epochs and size the decay schedule to that budget~\citep{hoffmann2022chinchilla}. We therefore shorten the budget to $2$, $4$, $5$, $6$, or $8$ epochs and decay the cosine schedule to one tenth of the peak within each (Table~\ref{tab:budget}, Figure~\ref{fig:mechanism}(c)). Taking the best of the five budgets for each condition, the \textsc{replace} run reaches a lower minimum than the \textsc{none} run in every seed. At $8$ epochs replacement lowers the minimum loss on both corpora. Within that budget the \textsc{none} run reaches its minimum near $4$ epochs and then rises, while the \textsc{replace} run keeps improving until about $6$ to $7$ epochs. The effect at the minimum loss grows with the budget, as \S\ref{sec:mechanism} predicts. At $2$ epochs both runs are still improving at their last step. No loss rise has appeared yet to offset the bias, so replacement raises the minimum loss on both corpora.

\begin{figure}[tb]
\centering
\includegraphics[width=\linewidth]{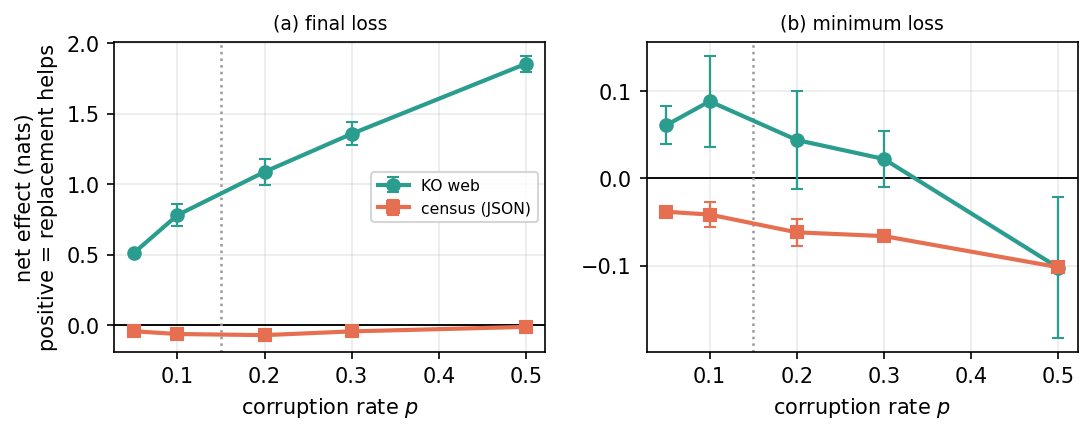}
\caption{Net effect of replacement in nats as the corruption rate $p$ varies, (a) on final loss and (b) at the minimum loss, positive when it helps. Error bars give the s.d. over three seeds, two for the census minimum below $p{=}0.5$; the dotted line marks the main rate $p{=}0.15$.}
\label{fig:psweep}
\end{figure}

\paragraph{The best corruption rate depends on the point of training.}
We sweep the corruption rate from $0.05$ to $0.50$ on two corpora at opposite ends (Figure~\ref{fig:psweep}, App.~\ref{app:seeds}), Korean web, where replacement clearly helps at $p{=}0.15$, and the census corpus, where it makes even the final loss worse. On final loss the benefit on Korean web grows with $p$, so heavier corruption looks better. At the minimum loss the benefit is largest at $p{=}0.10$, then shrinks and turns negative on average at $p{=}0.50$. Heavier corruption delays the loss rise more but also raises the bias, since $1-\alpha_{\tilde c}$ grows with $p$ in Eq.~\eqref{eq:alpha}. On the census corpus the final-loss effect stays slightly negative at every rate instead of growing with $p$, while the effect at the minimum loss falls steadily as $p$ rises. This is the pattern \S\ref{sec:mechanism} expects on highly predictable text such as the Markov source $t02$ (\S\ref{sec:mem}), where the bias is large and the benefit small; on the census corpus replacement removes little of the overfitting gap (Table~\ref{tab:seed}).

\paragraph{Robustness across model scale.}
The effect is not an artifact of one model size. Table~\ref{tab:scale} repeats the comparison at 134M for three corpora. Replacement removes a larger fraction of the overfitting gap at the smaller 134M scale than at 336M, and at 134M it also lowers the minimum loss more on both natural-language corpora. The census corpus is harmed at both sizes, its final loss rising by $0.05$ nats at 134M and $0.06$ nats at 336M (Table~\ref{app:tab:scaleseed}).

\section{Predicting the Sign on Real Corpora}
\label{sec:redundancy}

On the Markov source the entropy rate $H$ measures how predictable the source is, and as $H$ rises the sign of the effect on final loss turns from harm to help (\S\ref{sec:mem}). The entropy rate of a real corpus is not known. We therefore ask whether a corpus statistic can stand in for it and predict the sign.

Table~\ref{tab:nat} orders the seven structured corpora by redundancy, how much of the text repeats predictable patterns, from least to most. Replacement lowers the final loss most on chess and code, has smaller effects on the two SQL forms and on the system-log and road-safety corpora, and raises it on the census corpus. On this corpus the baseline overfits, rising from $0.26$ to $0.42$ nats, yet the replacement run ends higher (Figure~\ref{fig:curves}(f)). MI is the statistic closest to \S\ref{sec:mechanism}, because it averages over contexts the distance between $P_{\tilde c}$ and $\pi$ that sets the bias (App.~\ref{app:alpha}). As MI rises the benefit broadly shrinks, which agrees with \S\ref{sec:mechanism}. MI does not separate the sign, however. The census corpus, the system-log corpus and the road-safety corpus lie within $0.02$ of each other in MI, yet only the census corpus is harmed, and replacement leaves the minimum loss of the other two nearly unchanged (Table~\ref{tab:seed}).

Both helped corpora have more training tokens than the census corpus, so at the fixed step budget they train for \emph{fewer} epochs. More epochs let the loss rise more, leaving more for replacement to remove. If repetition decided the sign, replacement would help the census corpus more than these two, yet it harms it. The difference in training tokens therefore does not explain the harm.

\begin{table}[tb]
\centering
\small
\resizebox{\columnwidth}{!}{%
\begin{tabular}{llcccc}
\toprule
Corpus & Label & gzip & MI & $\Delta L_{\text{final}}$ & $\Delta L_{\text{min}}$ \\
\midrule
chess & S & $0.386$ & $0.42$ & $+0.29{\pm}0.04^{*}$ & $+0.03{\pm}0.01^{*}$ \\
code & S & $0.199$ & $0.40$ & $+0.24{\pm}0.04^{*}$ & $-0.06{\pm}0.03$ \\
SQL-comb & S, P & $0.170$ & $0.62$ & $+0.04{\pm}0.04$ & $+0.06{\pm}0.05$ \\
SQL-ctx & S, P & $0.125$ & $0.69$ & $+0.05{\pm}0.09$ & $+0.13{\pm}0.02^{*}$ \\
road-safety (JSON)$^{\dagger}$ & S, P & $0.068$ & $0.74$ & $+0.04{\pm}0.00^{*}$ & $+0.00{\pm}0.00$ \\
loghub$^{\dagger}$ & S, P & $0.063$ & $0.72$ & $+0.07{\pm}0.01$ & $+0.02{\pm}0.02$ \\
census (JSON) & S, P & $0.049$ & $0.73$ & $-0.06{\pm}0.02^{*}$ & $-0.06{\pm}0.02$ \\
\bottomrule
\end{tabular}}
\caption{Change in the final and minimum loss from replacement, in nats, mean${\pm}$s.d. over three seeds, positive when replacement helps; $^{*}$ is $p<0.05$. S is structured and P program-generated (\S\ref{sec:setup}). $^{\dagger}$Two seeds.}
\label{tab:nat}
\end{table}

Program-generated origin and higher-order conditional entropy do not mark the census corpus either (App.~\ref{app:cond}). Under replacement the minimum loss of the census corpus rises by $0.06$ nats, though not significantly, whereas on both helped corpora it is nearly unchanged. A low minimum does not mark it either, since the helped road-safety corpus has a lower one (Table~\ref{tab:seed}). Among the statistics we test, only the gzip ratio sets the census corpus apart.

\begin{figure}[t]
\centering
\includegraphics[width=\linewidth]{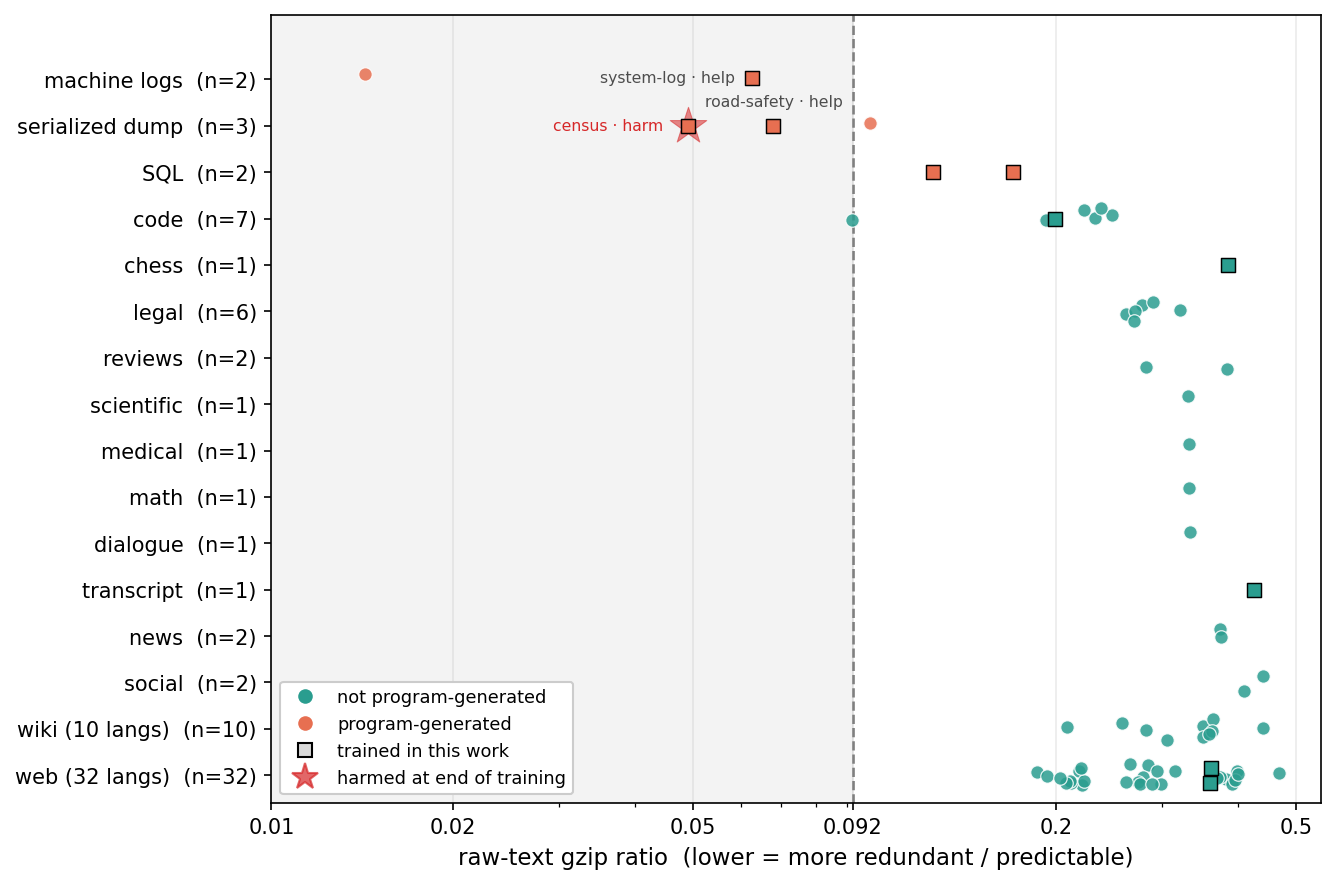}
\caption{Raw-text gzip ratio of the surveyed corpora and of the ten corpora we train on, by category.}
\label{fig:survey}
\end{figure}

In gzip ratio the census corpus is the most redundant corpus we train on, and the helped system-log and road-safety corpora sit just above it, at $0.063$ and $0.068$. One harmed corpus cannot show whether a gzip threshold between these values separates the sign. It can still show at what redundancy the harm occurs, so we ask how often real text is as redundant as the census corpus. We measure raw-text gzip on $64$ corpora that are not program-generated, spanning more than $30$ languages and 11 domains (Figure~\ref{fig:survey}, App.~\ref{app:survey}). Every one sits well above the census corpus, the most redundant being Go source code at $0.092$. Only program-generated text falls below that value, and the one harm we observe lies there. Every corpus we train on that is not program-generated is helped by the end of the budget (Table~\ref{tab:seed}).

\section{Conclusion}
\label{sec:conclusion}

We decompose validation loss into APPROX, EST, and MEM terms on a source of known entropy, which shows what random-token replacement removes. The decomposition needs only validation losses, so it applies to any augmentation. The benefit of replacement is a smaller loss rise after the minimum, and its cost is a bias that is present from the start. The point at which training stops therefore decides the size and sign of the effect. On final loss this benefit dominates, and learning-rate decay does not replace it. At the minimum loss the bias is exposed and the gain is small, though with a schedule sized to the budget replacement still reaches a lower best minimum. On the controlled source the sign on final loss also turns from harm to help as predictability falls. On real corpora replacement helps every corpus we test that is not program-generated, and harms only the census corpus, which is more redundant than all of them.

\section*{Limitations}

\begin{itemize}
\itemsep-0.2em
\item \textbf{Model scale.} Our study spans two model scales, 134M and 336M; the benefit is present at both but diminishes as scale grows, and we do not claim it persists at substantially larger scale.
\item \textbf{Harm on real corpora.} On final loss, replacement harms one real corpus we test, so we observe the predictability threshold only on the controlled source. Locating it on real corpora needs more corpora that cross it, and building such a set is our next step.
\item \textbf{Cross-family comparisons and controls.} Several structured corpora hold fewer training tokens than the web and transcript corpora, so under the fixed step budget they train for more epochs. Cross-family sign comparisons therefore mix predictability with repetition dose; the equal-token web and transcript corpora and the controlled Markov source are the clean comparisons. An English-prose control under the same recipe would tie our runs to the published English effect, and it is the next control we would add.
\item \textbf{Corruption objective and rate.} We sweep the corruption rate on two corpora, and the remaining experiments use $p{=}0.15$. We size the decayed schedule to five budgets from $2$ to $8$ epochs on web text, and to five from $4$ to $66$ on the controlled source. Dropout and other explicit regularizers, the masking and infilling variants, rates beyond $p{=}0.5$, and decayed budgets longer than $8$ epochs on real corpora are the sweeps we would run next.
\item \textbf{Languages.} Both languages we test use non-Latin scripts but have abundant data overall. Under-resourced languages and writing systems with higher tokenizer fertility are the setting we would test next.
\item \textbf{Controlled source and metric.} The controlled analysis uses an order-1 Markov source; higher-order sources would test the mechanism further. Reported effects are based on validation loss, which suits the pretraining regime we study, and they may not transfer to task performance.
\end{itemize}

\section*{Ethical Considerations}
This work studies a training-time regularization method for small language models trained from scratch. All corpora are publicly available (\S\ref{sec:setup}, App.~\ref{app:repro}) and are used rather than re-released; the YODAS speech transcripts are public automatic captions, and we perform no extraction of personally identifying information. The computational footprint is modest (about three GPU-hours per run on a single GPU, App.~\ref{app:repro}), limiting environmental cost. The method targets pretraining loss with no deployment or downstream claims, so we identify no significant additional ethical, societal, or dual-use risks.

\bibliography{references}

\appendix

\section{Training and Reproducibility Details}
\label{app:repro}

\paragraph{Model.} The 336M model is a decoder-only Transformer with 24 layers, hidden size 1024, 16 attention heads in 4 key-value groups, and feed-forward width 2816; the 134M variant has 14 layers, hidden size 768, 12 heads in 2 key-value groups, and feed-forward width 2048. On the Markov source the vocabulary is 1025 rather than 64k, which lowers the parameter count to 271.6M at the 336M configuration. All runs use bfloat16 and gradient checkpointing.

\paragraph{Recipe.} Constant learning rate $3\times10^{-4}$ with 300 warmup steps, AdamW with $\beta=(0.9,0.95)$ and weight decay $0.1$, gradient clipping at $1.0$, batch size 8, sequence length 2048, 10{,}000 steps, $8{,}000$ on the Markov source, and a shared 64k SentencePiece tokenizer. Replacement independently replaces each input token with a uniform random vocabulary token at rate $p{=}0.15$, leaving prediction targets intact. The learning-rate comparison of \S\ref{sec:safe} uses a cosine schedule from the same peak to $3\times10^{-5}$ over the $10{,}000$ steps, with the same warmup. These runs and the constant-rate runs they are compared with were trained on an H100 GPU. The H100 constant-rate runs reproduce the main-experiment runs within $0.04$ nats at both points of training. The budget-sized schedules of Table~\ref{tab:budget} run for $900$ to $3{,}600$ steps, which is $2$, $4$, $5$, $6$, or $8$ epochs of a $7.4$M-token corpus, with the same warmup, cosine decay to $3\times10^{-5}$ at the last step, and an evaluation every $150$ steps. The abundant-data runs at $t02$ were repeated as a pair on an H100 GPU, where replacement raises $L_{\text{abund}}$ by $0.048$ nats.

\paragraph{Data split.} For every corpus we draw the training tokens and a disjoint validation slice from the same source. The $7.4$M-token corpora have exactly $7{,}383{,}561$ training tokens and a separate $0.5$M-token validation slice. All $L_{\text{min}}$ and $L_{\text{final}}$ values in the paper are computed on this held-out slice, which the model never sees in training; the Markov source's split is given below.

\paragraph{Compute and availability.} Every run trains a single 134M or 336M model on one L4 or L40S GPU in about three hours, or on one H100 GPU in under an hour, with no multi-GPU or multi-node parallelism. All corpora in Table~\ref{app:tab:corpora} are public, and our training code and configuration files are available at \url{https://github.com/wonsupshin-flitto/random-token-replacement-repeated-data}, so every result can be reproduced.

\paragraph{Markov source.} The controlled source is an order-1 Markov chain over $V{=}1024$ states. The model vocabulary has $V{+}1$ ids. The extra id is reserved for masking and never occurs in the data, and replacement draws uniformly from all $V{+}1$ ids. Each row $P_i$ of the transition matrix $P$, with entries $P_{ij}$, is ${\rm softmax}(g_i/T)$ with independent standard-normal logits $g_i$ and a fixed seed, so the chain is reproducible, and the temperature $T$ sets predictability: small $T$ gives near-deterministic rows, while large $T$ makes them uniform with $H{\to}\log V$. Writing $\pi$ for the stationary distribution, $\pi P=\pi$, the entropy rate has the closed form $H=-\sum_i \pi_i \sum_j P_{ij}\ln P_{ij}$~\citep{cover2006elements}, which serves as the lower bound in \S\ref{sec:mem}. Training and validation sequences are disjoint samples from the same fixed chain, $2$M and $0.2$M tokens, and the Markov runs train for $8{,}000$ steps, about $65$ epochs.

\paragraph{Abundant-data reference.} The approximation term APPROX${=}L_{\text{abund}}{-}H$ uses an \emph{abundant-data} run, with the same model and recipe but a much larger draw from the same Markov source ($1.4\times10^{8}$ tokens versus the $2\times10^{6}$-token data-constrained corpus), so tokens repeat little and memorization does not arise. Its loss $L_{\text{abund}}$ is $1.686$, $5.085$, and $6.878$ nats at $t02$, $t05$, and $t20$ for the baseline, and $1.727$, $5.139$, and $6.901$ under replacement. In every abundant-data run, validation loss rises by at most $0.004$ nats after its minimum, against $2.0$ to $5.4$ nats on average in the data-constrained runs at the same temperatures.

\paragraph{Corpora.} Table~\ref{app:tab:corpora} lists sources and training-token counts. The two SQL forms are labeled program-generated because their CREATE TABLE contexts were generated with SQLGlot from Spider and WikiSQL queries.

\begin{table}[h]
\centering\small
\begin{tabular}{lll}
\toprule
Corpus & Src & Train tok. \\
\midrule
KO web & FineWeb-2 & $7.4$M \\
JA web & FineWeb-2 & $7.4$M \\
KO transcript & YODAS & $7.4$M \\
code & The Stack & $7.4$M \\
chess & move sequences & $5.2$M \\
SQL-comb & SQL-create-context & $5.0$M \\
SQL-ctx & SQL-create-context & $2.3$M \\
census (JSON) & Adult census & $5.5$M \\
road-safety (JSON) & road-safety table & $7.4$M \\
loghub & LogHub & $7.4$M \\
Markov (constr.) & order-1 chain & $2.0$M \\
Markov (abund.) & order-1 chain & $140$M \\
\bottomrule
\end{tabular}
\caption{Corpora, sources, and training-token counts.}
\label{app:tab:corpora}
\end{table}

\section{Detailed Results}

\subsection{Loss Decomposition on the Markov Source}
\label{app:decomp}

Table~\ref{tab:decomp} lists the exact three-seed decomposition values plotted in Figure~\ref{fig:memdecomp}: the approximation (APPROX), estimation (EST), and memorization (MEM) terms of Eq.~\eqref{eq:decomp} at three source temperatures.

\begin{table}[h]
\centering
\small
\resizebox{\columnwidth}{!}{%
\begin{tabular}{llcccc}
\toprule
Src & Cond & $H$ & APPROX & EST & MEM \\
\midrule
$t02$ & none & $1.646$ & $0.040$ & $0.102$ & $2.29{\pm}0.01$ \\
$t02$ & replace & $1.646$ & $0.081$ & $0.177$ & $2.41{\pm}0.08$ \\
$t05$ & none & $5.019$ & $0.066$ & $0.212$ & $4.78{\pm}0.04$ \\
$t05$ & replace & $5.019$ & $0.120$ & $0.157$ & $\mathbf{2.02{\pm}0.00}$ \\
$t20$ & none & $6.806$ & $0.072$ & $0.057$ & $5.39{\pm}0.15$ \\
$t20$ & replace & $6.806$ & $0.095$ & $0.033$ & $\mathbf{2.05{\pm}0.39}$ \\
\bottomrule
\end{tabular}}
\caption{Loss decomposition on the Markov source in nats, three-seed means, with MEM as mean${\pm}$s.d.; APPROX and EST vary by less than $0.01$ nats across seeds.}
\label{tab:decomp}
\end{table}

\subsection{Redundancy Survey}
\label{app:survey}

Table~\ref{app:tab:survey} gives the full redundancy survey behind Figure~\ref{fig:survey}: $64$ corpora that are not program-generated, across more than $30$ languages and 11 domains, plus program-generated text. No corpus that is not program-generated falls below raw-text gzip $0.092$, the value of Go source code from CodeSearchNet~\citep{husain2019codesearchnet}, the most redundant one; among the surveyed corpora only the system logs do. The two JSON serializations we train on, at gzip $0.049$ and $0.068$, also lie below it but were not part of the survey; the survey's nearby program-generated entry is a distinct serialized dump at gzip $0.098$.

\begin{table}[h]
\centering\small
\begin{tabular}{lccc}
\toprule
Category & $n$ & gzip & MI/H1 \\
\midrule
web (FineWeb-2) & 32 & $0.186$--$0.468$ & $0.32$--$0.52$ \\
Wikipedia & 10 & $0.209$--$0.441$ & $0.33$--$0.48$ \\
legal & 6 & $0.262$--$0.322$ & $0.42$--$0.49$ \\
code & 6 & $0.092$--$0.248$ & $0.42$--$0.64$ \\
reviews & 2 & $0.282$--$0.384$ & $0.41$--$0.45$ \\
scientific & 1 & $0.331$ & $0.42$ \\
medical & 1 & $0.332$ & $0.44$ \\
math & 1 & $0.333$ & $0.42$ \\
dialogue & 1 & $0.334$ & $0.42$ \\
news & 2 & $0.375$--$0.376$ & $0.40$--$0.47$ \\
social & 2 & $0.411$--$0.441$ & $0.50$--$0.53$ \\
\midrule
serialized dump & 1 & $0.098$ & $0.66$ \\
machine logs & 2 & $0.014$--$0.063$ & $0.71$--$0.81$ \\
\bottomrule
\end{tabular}
\caption{Redundancy survey by category, ranges of raw-text gzip ratio and normalized bigram MI.}
\label{app:tab:survey}
\end{table}

\subsection{Predictability Scalars Do Not Separate the Census Corpus}
\label{app:cond}

Table~\ref{app:tab:cond} reports higher-order conditional entropies, measured on $2$M-token samples with the shared tokenizer. No conditional-entropy scalar separates the census corpus from the helped corpora: a helped web corpus even has a lower order-3 conditional entropy than the census corpus, an artifact of data sparsity in low-repetition text. Among the corpora in Table~\ref{app:tab:cond}, the census corpus also has the lowest baseline minimum loss, at $0.26$ nats against $0.93$ for loghub. This does not separate it either. The helped road-safety corpus, not in the table, has a lower minimum still, at $0.17$ nats (Table~\ref{tab:seed}).

\begin{table}[h]
\centering\small
\resizebox{\columnwidth}{!}{%
\begin{tabular}{lcccccc}
\toprule
Corpus & $H_1$ & $H_{c1}$ & $H_{c2}$ & $H_{c3}$ & MI$_1/H_1$ & $L_{\text{min}}$ \\
\midrule
census (harm) & $5.76$ & $1.56$ & $0.85$ & $0.57$ & $0.73$ & $0.26$ \\
loghub (help) & $6.58$ & $1.89$ & $1.00$ & $0.69$ & $0.72$ & $0.93$ \\
SQL-ctx (noise) & $6.12$ & $1.88$ & $1.26$ & $1.07$ & $0.69$ & $2.40$ \\
code (help) & $8.36$ & $5.01$ & $2.65$ & $1.43$ & $0.40$ & $1.86$ \\
KO web (help) & $11.65$ & $6.23$ & $1.86$ & $0.50$ & $0.47$ & $4.91$ \\
\bottomrule
\end{tabular}}
\caption{Unigram entropy $H_1$ and order-1 to order-3 conditional entropies $H_{c1},H_{c2},H_{c3}$ in bits on $2$M-token samples, normalized bigram MI, MI$_1/H_1$, and the baseline's minimum loss $L_{\text{min}}$ in nats. The row label gives the sign of the final-loss effect; $L_{\text{min}}$ uses three seeds, two for loghub.}
\label{app:tab:cond}
\end{table}

\subsection{Per-Seed Results}
\label{app:seeds}

Table~\ref{app:tab:natseed} reports the per-seed minimum and final losses behind the web-text results, and Table~\ref{app:tab:doseseed} the per-seed effects behind the Markov sweep of Figure~\ref{fig:mechanism}(b). Tables~\ref{app:tab:psweepseed} and~\ref{app:tab:psweepminseed} report the per-seed net effects behind the two panels of the corruption-rate sweep of Figure~\ref{fig:psweep}, Tables~\ref{app:tab:scaleseed} and~\ref{app:tab:budgetseed} report those behind the two-scale comparison and the best-budget comparison of \S\ref{sec:safe}. We assess significance with a one-sample $t$-test on the three-seed effects, two-sided with df${=}2$, and mark $p<0.05$ with $^{*}$ in the tables. The two SQL forms, described as within noise in \S\ref{sec:safe}, have $p>0.2$.
\begin{table}[h]
\centering\small
\begin{tabular}{llcccc}
\toprule
Corpus & Cond & & s1 & s2 & s3 \\
\midrule
KO web & none & $L_{\text{min}}$ & $4.86$ & $5.01$ & $4.87$ \\
 & & $L_{\text{final}}$ & $7.29$ & $7.53$ & $7.32$ \\
KO web & replace & $L_{\text{min}}$ & $4.87$ & $4.85$ & $4.84$ \\
 & & $L_{\text{final}}$ & $6.44$ & $6.43$ & $6.47$ \\
JA web & none & $L_{\text{min}}$ & $4.75$ & $4.65$ & $4.71$ \\
 & & $L_{\text{final}}$ & $7.47$ & $7.55$ & $7.48$ \\
JA web & replace & $L_{\text{min}}$ & $4.65$ & $4.57$ & $4.62$ \\
 & & $L_{\text{final}}$ & $6.40$ & $6.40$ & $6.28$ \\
\bottomrule
\end{tabular}
\caption{Per-seed minimum loss and final loss in nats for the two web corpora at 336M.}
\label{app:tab:natseed}
\end{table}

\begin{table}[h]
\centering\small
\begin{tabular}{lcccc}
\toprule
Src & net s1 & net s2 & net s3 & mean${\pm}$s.d. \\
\midrule
$t02$ & $-0.35$ & $-0.19$ & $-0.17$ & $-0.23{\pm}0.08$ \\
$t03$ & $+1.00$ & $+1.14$ & $+0.95$ & $+1.03{\pm}0.08$ \\
$t05$ & $+2.72$ & $+2.75$ & $+2.82$ & $+2.76{\pm}0.04$ \\
$t08$ & $+2.69$ & $+2.37$ & $+1.83$ & $+2.29{\pm}0.36$ \\
$t20$ & $+3.02$ & $+3.62$ & $+3.40$ & $+3.35{\pm}0.25$ \\
\bottomrule
\end{tabular}
\caption{Per-seed change in the final loss, $\Delta L_{\text{final}}$ in nats, across the Markov temperatures, with the three-seed mean${\pm}$s.d.}
\label{app:tab:doseseed}
\end{table}

\begin{table}[h]
\centering\small
\begin{tabular}{lcccc}
\toprule
$p$ & s1 & s2 & s3 & mean${\pm}$s.d. \\
\midrule
\multicolumn{5}{l}{\emph{KO web}} \\
$0.05$ & $0.51$ & $0.52$ & $0.50$ & $0.51{\pm}0.01$ \\
$0.10$ & $0.74$ & $0.87$ & $0.73$ & $0.78{\pm}0.08$ \\
$0.20$ & $1.04$ & $1.19$ & $1.03$ & $1.09{\pm}0.09$ \\
$0.30$ & $1.29$ & $1.45$ & $1.34$ & $1.36{\pm}0.08$ \\
$0.50$ & $1.83$ & $1.92$ & $1.82$ & $1.85{\pm}0.06$ \\
\midrule
\multicolumn{5}{l}{\emph{census corpus}} \\
$0.05$ & $-0.03$ & $-0.06$ & $-0.04$ & $-0.04{\pm}0.01$ \\
$0.10$ & $-0.05$ & $-0.06$ & $-0.08$ & $-0.06{\pm}0.01$ \\
$0.20$ & $-0.09$ & $-0.07$ & $-0.05$ & $-0.07{\pm}0.02$ \\
$0.30$ & $-0.06$ & $-0.05$ & $-0.03$ & $-0.04{\pm}0.02$ \\
$0.50$ & $-0.01$ & $0.00$ & $-0.03$ & $-0.01{\pm}0.02$ \\
\bottomrule
\end{tabular}
\caption{Per-seed change in the final loss, $\Delta L_{\text{final}}$ in nats, across the corruption rates of Figure~\ref{fig:psweep}(a) on Korean web and the census corpus, with the three-seed mean${\pm}$s.d.}
\label{app:tab:psweepseed}
\end{table}

\begin{table}[h]
\centering\small
\begin{tabular}{lcccc}
\toprule
$p$ & s1 & s2 & s3 & mean${\pm}$s.d. \\
\midrule
\multicolumn{5}{l}{\emph{KO web}} \\
$0.05$ & $+0.051$ & $+0.091$ & $+0.041$ & $+0.061{\pm}0.022$ \\
$0.10$ & $+0.063$ & $+0.161$ & $+0.042$ & $+0.089{\pm}0.052$ \\
$0.20$ & $-0.017$ & $+0.119$ & $+0.031$ & $+0.044{\pm}0.057$ \\
$0.30$ & $+0.006$ & $+0.067$ & $-0.006$ & $+0.022{\pm}0.032$ \\
$0.50$ & $-0.105$ & $-0.002$ & $-0.200$ & $-0.102{\pm}0.081$ \\
\midrule
\multicolumn{5}{l}{\emph{census corpus}} \\
$0.05$ & $-0.039$ & $-0.038$ & -- & $-0.038{\pm}0.001$ \\
$0.10$ & $-0.056$ & $-0.027$ & -- & $-0.041{\pm}0.015$ \\
$0.20$ & $-0.077$ & $-0.046$ & -- & $-0.062{\pm}0.015$ \\
$0.30$ & $-0.070$ & $-0.062$ & -- & $-0.066{\pm}0.004$ \\
$0.50$ & $-0.106$ & $-0.098$ & $-0.102$ & $-0.102{\pm}0.003$ \\
\bottomrule
\end{tabular}
\caption{Per-seed change in the minimum loss, $\Delta L_{\text{min}}$ in nats, across the corruption rates of Figure~\ref{fig:psweep}(b). Two seeds are available for the census corpus below $p{=}0.5$.}
\label{app:tab:psweepminseed}
\end{table}

\begin{table}[h]
\centering\small
\begin{tabular}{lcccc}
\toprule
Scale & s1 & s2 & s3 & mean${\pm}$s.d. \\
\midrule
$134$M & $-0.06$ & $-0.03$ & $-0.06$ & $-0.05{\pm}0.01$ \\
$336$M & $-0.05$ & $-0.04$ & $-0.08$ & $-0.06{\pm}0.02$ \\
\bottomrule
\end{tabular}
\caption{Per-seed change in the final loss from replacement on the census corpus, $\Delta L_{\text{final}}$ in nats, at both model scales; negative means replacement harms. The $336$M row is the \emph{census (JSON)} row of Table~\ref{tab:nat}.}
\label{app:tab:scaleseed}
\end{table}

\begin{table}[h]
\centering\small
\resizebox{\columnwidth}{!}{%
\begin{tabular}{lcccc}
\toprule
Corpus & s1 & s2 & s3 & mean${\pm}$s.d. \\
\midrule
KO web & $+0.052$ & $+0.288$ & $+0.148$ & $+0.163{\pm}0.097$ \\
JA web & $+0.089$ & $+0.065$ & $+0.082$ & $+0.079{\pm}0.010$ \\
\bottomrule
\end{tabular}}
\caption{Per-seed change in the minimum loss, $\Delta L_{\text{min}}$ in nats, behind the best-budget row of Table~\ref{tab:budget}. Each condition uses its minimum loss at its own best budget within each seed.}
\label{app:tab:budgetseed}
\end{table}

\section{Supplementary Experiments}

\subsection{Memorization and Shrinkage Probes on the Markov Source}
\label{app:memprobe}

To check that MEM arises from memorization, we measure each run's loss on clean training and validation windows of the same length. Every run fits its training windows to a loss far below the entropy rate $H$, the lowest loss any model can reach on held-out windows from the source (Table~\ref{app:tab:memprobe}). All runs therefore memorize their training sequences. Replacement leaves the training loss about the same or lower, and lowers the validation loss where it helps. We also measure verbatim extraction~\citep{carlini2023quantifying}, the share of 50-token spans after at least 1{,}024 tokens of context that greedy decoding reproduces exactly. Replacement raises this share on training windows at every temperature, while it lowers MEM at $t05$ and $t20$ (\S\ref{sec:mem}), and the share is zero on validation windows in every run. MEM therefore does not measure how much the model memorizes. It measures the rise in held-out loss that accompanies memorization.

\begin{table}[h]
\centering\small
\resizebox{\columnwidth}{!}{%
\begin{tabular}{llcccccc}
\toprule
Source & Cond & $H$ & Train & Val & Extr. & $\mathrm{KL}(Q_{\theta}\Vert P_{\text{true}})$ & $\mathrm{KL}(Q_{\theta}\Vert \pi)$ \\
\midrule
$t02$ & none & $1.65$ & $0.14$ & $4.06$ & $0.69$ & $0.15$ & $5.31$ \\
$t02$ & replace & & $0.04$ & $4.40$ & $0.90$ & $2.55$ & $3.57$ \\
$t05$ & none & $5.02$ & $0.30$ & $10.02$ & $0.55$ & $0.56$ & $2.52$ \\
$t05$ & replace & & $0.30$ & $7.32$ & $0.79$ & $0.66$ & $1.24$ \\
$t20$ & none & $6.81$ & $0.83$ & $12.47$ & $0.40$ & $0.60$ & $0.49$ \\
$t20$ & replace & & $0.70$ & $9.48$ & $0.56$ & $0.30$ & $0.27$ \\
\bottomrule
\end{tabular}}
\caption{Clean training and validation loss in nats, the share of 50-token training spans reproduced verbatim (Extr.), and $\pi$-weighted mean $\mathrm{KL}(Q_{\theta}\Vert\cdot)$ to the true transition $P_{\text{true}}$ and the marginal $\pi$, on the Markov source. Each loss averages 160 windows of 2{,}048 tokens.}
\label{app:tab:memprobe}
\end{table}

We also compare each model's learned next-token distribution $Q_{\theta}(\cdot\mid c)$, aggregated over validation positions by the preceding context $c$, with the true transition $P_{\text{true}}$ and the corpus marginal $\pi$. The table reports, for each reference $R$, the $\pi$-weighted mean $\sum_c \pi(c)\,\mathrm{KL}(Q_{\theta}(\cdot\mid c)\Vert R(\cdot\mid c))$, with $R(\cdot\mid c)=\pi$ for the marginal. Replacement moves $Q_{\theta}$ toward $\pi$ at every temperature, the shrinkage of \S\ref{sec:mechanism}. Its effect on the distance to $P_{\text{true}}$ follows predictability. The distance rises sharply at the most predictable source $t02$, barely changes at $t05$, and falls at $t20$, which agrees with the bias that \S\ref{sec:mechanism} predicts.

\subsection{Budget-Sized Schedules on the Markov Source}
\label{app:mkbudget}

Table~\ref{tab:budget} varies the training budget on web text with the cosine
schedule decayed within each budget. We repeat that design on the Markov source,
using the same chains as \S\ref{sec:mem} at three temperatures.
Budgets run from $500$ to $8{,}000$ steps, which is $4$ to $66$ epochs, with
three seeds per cell, for $90$ runs in total.

At $t02$ replacement raises the minimum at every budget, which agrees with the shrinkage
bias of \S\ref{sec:mechanism} (Table~\ref{app:tab:mkbudget}). At $t20$ it leaves
the minimum unchanged at every budget. The largest improvement in any cell is
$0.023$ nats, at $t05$ under a $16$-epoch budget.

\begin{table}[h]
\centering
\small
\resizebox{\columnwidth}{!}{%
\begin{tabular}{lccc}
\toprule
Budget & $t02$ & $t05$ & $t20$ \\
\midrule
$4$ epochs & $-0.149{\pm}0.001^{*}$ & $-0.042{\pm}0.001^{*}$ & $+0.001{\pm}0.003$ \\
$8$ epochs & $-0.106{\pm}0.005^{*}$ & $-0.028{\pm}0.000^{*}$ & $+0.000{\pm}0.000$ \\
$16$ epochs & $-0.113{\pm}0.007^{*}$ & $+0.023{\pm}0.002^{*}$ & $-0.001{\pm}0.003$ \\
$33$ epochs & $-0.117{\pm}0.009^{*}$ & $+0.003{\pm}0.004$ & $+0.002{\pm}0.001$ \\
$66$ epochs & $-0.119{\pm}0.009^{*}$ & $-0.001{\pm}0.006$ & $+0.001{\pm}0.001$ \\
\bottomrule
\end{tabular}}
\caption{Change in the minimum loss on the Markov source,
$\Delta L_{\text{min}}$ in nats, with the cosine schedule
decayed within each budget, mean${\pm}$s.d. over three seeds; $^{*}$ is
$p<0.05$.}
\label{app:tab:mkbudget}
\end{table}

\subsection{Shrinkage Weight on Real Corpora}
\label{app:alpha}

Eq.~\eqref{eq:shrink} in \S\ref{sec:mechanism} writes the optimal prediction after an observed context token
$\tilde c$ as $\alpha_{\tilde c}P_{\tilde c}+(1-\alpha_{\tilde c})\pi$, where $P_{\tilde c}$ is the true conditional, $\pi$ the corpus marginal, and $\alpha_{\tilde c}$ the posterior probability that $\tilde c$ is clean. Averaged over
observed positions, $1-\alpha$ equals the corruption rate $p$ on every corpus, because
by Eq.~\eqref{eq:alpha}, $(1-\alpha_{\tilde c})$ times the probability of observing $\tilde c$ is $p\,q(\tilde c)$, where $q$ is the noise distribution and sums to one.
What differs between corpora is where that weight falls. We report two quantities, both
computed from the training tokens with no model involved.

The first, \emph{out-of-corpus} in Table~\ref{app:tab:alpha}, is how often the uniform noise draws a token that never occurs in the corpus. Such a token
is recognizable as noise, and the model can fall back to the marginal at that position alone, so no clean context is shrunk. The
second is the shrinkage that falls on clean contexts, $w=\sum_c \pi(c)(1-\alpha_c)$. That
is the part that costs accuracy, since it dilutes a context that was in fact clean.
The bias also grows with how far the true conditional departs from the marginal. The expectation of that distance over contexts is the bigram mutual information, MI, used here in nats rather than normalized as in \S\ref{sec:setup}. We therefore report the product $w\,$MI as a predicted bias. Table~\ref{app:tab:alpha} gives all three at $p{=}0.15$.

On the Markov source almost no noise is recognizable, so shrinkage falls on every
prediction and $w$ is $0.137$ to $0.150$, close to $p$. On the real corpora $34$ to $99.7\%$ of noise
draws are out-of-corpus and $w$ is $0.001$ to $0.058$. Across the three Markov
temperatures $w\,$MI tracks the rise in the minimum loss, from $0.69$ at $t02$, where the
minimum rises by $0.12$ nats, to $0.06$ at $t20$, where it is unchanged. Across the real
corpora $w\,$MI does not separate the harmed corpus from the helped ones. The harmed census corpus and the helped loghub and road-safety corpora sit within $0.001$ of each other on $w\,$MI. This matches \S\ref{sec:redundancy}, where mutual information does not set the census corpus apart either.

\begin{table}[h]
\centering
\small
\resizebox{\columnwidth}{!}{%
\begin{tabular}{lccccc}
\toprule
Corpus & out-of-corpus (\%) & $w$ & MI & $w\,$MI & $\Delta L_{\text{min}}$ \\
\midrule
KO web & $33.7$ & $0.058$ & $3.34$ & $0.194$ & $+0.06$ \\
JA web & $63.6$ & $0.036$ & $2.80$ & $0.101$ & $+0.09$ \\
KO transcript & $55.6$ & $0.038$ & $3.35$ & $0.128$ & $+0.06$ \\
code & $81.8$ & $0.016$ & $2.11$ & $0.033$ & $-0.06$ \\
chess & $99.7$ & $0.000$ & $1.64$ & $0.001$ & $+0.03$ \\
SQL-comb & $87.8$ & $0.010$ & $3.09$ & $0.031$ & $+0.06$ \\
SQL-ctx & $93.9$ & $0.006$ & $2.94$ & $0.018$ & $+0.13$ \\
census & $99.0$ & $0.001$ & $2.91$ & $0.004$ & $-0.06$ \\
loghub & $98.6$ & $0.002$ & $2.89$ & $0.005$ & $+0.02$ \\
road-safety & $99.2$ & $0.001$ & $3.18$ & $0.004$ & $+0.00$ \\
Markov t02 & $0.1$ & $0.137$ & $5.01$ & $0.689$ & $-0.12$ \\
Markov t05 & $0.1$ & $0.149$ & $2.08$ & $0.311$ & $+0.00$ \\
Markov t20 & $0.1$ & $0.150$ & $0.42$ & $0.063$ & $+0.00$ \\
\bottomrule
\end{tabular}}
\caption{Shrinkage measured from the training tokens at $p{=}0.15$. \emph{out-of-corpus}
is the chance that the uniform noise draws a token that never occurs in the corpus; $w$ is the shrinkage
falling on clean contexts; MI is the bigram mutual information in nats. $\Delta L_{\text{min}}$
repeats the measured effect at the minimum loss from Table~\ref{tab:seed} and
\S\ref{sec:mem}.}
\label{app:tab:alpha}
\end{table}

\subsection{Minimum Loss and Early Stopping}
\label{app:es}

Our minimum loss is read from the full curve after training, whereas early stopping halts once validation loss stops improving for a set number of evaluations and keeps the best checkpoint seen so far. The two can differ when the loss rises for a few evaluations and then falls again before its true minimum. We simulate early stopping on the 174 runs whose minimum loss the paper reports, evaluated every 250 or 300 steps. With a patience of one evaluation it misses the minimum on 79 runs, by up to $0.65$ nats. With a patience of three it misses on 10, by up to $0.09$, and with five on 2, by at most $0.01$. The runs it misses at low patience are mostly \textsc{replace} runs, 58 of the 79 at a patience of one, because their loss descends more slowly. The comparison in \S\ref{sec:safe} therefore needs a patience of about five evaluations, which is 1{,}500 steps or about three epochs on the $7.4$M-token corpora.

\subsection{Weight Decay}
\label{app:wd}

To check that replacement is not a proxy for stronger explicit regularization, we raise the weight decay from $0.1$ to $0.5$ and compare with and without replacement. On Korean web and code, two seeds each, replacement still removes $46\%$ and $39\%$ of the overfitting gap, and at the minimum loss it lowers the loss on Korean web and leaves it unchanged on code. It therefore acts on a part of the gap that weight decay does not remove.

\subsection{Epoch Sweep on Repeated Data}
\label{app:wall}

\begin{figure}[t]
\centering
\includegraphics[width=\linewidth]{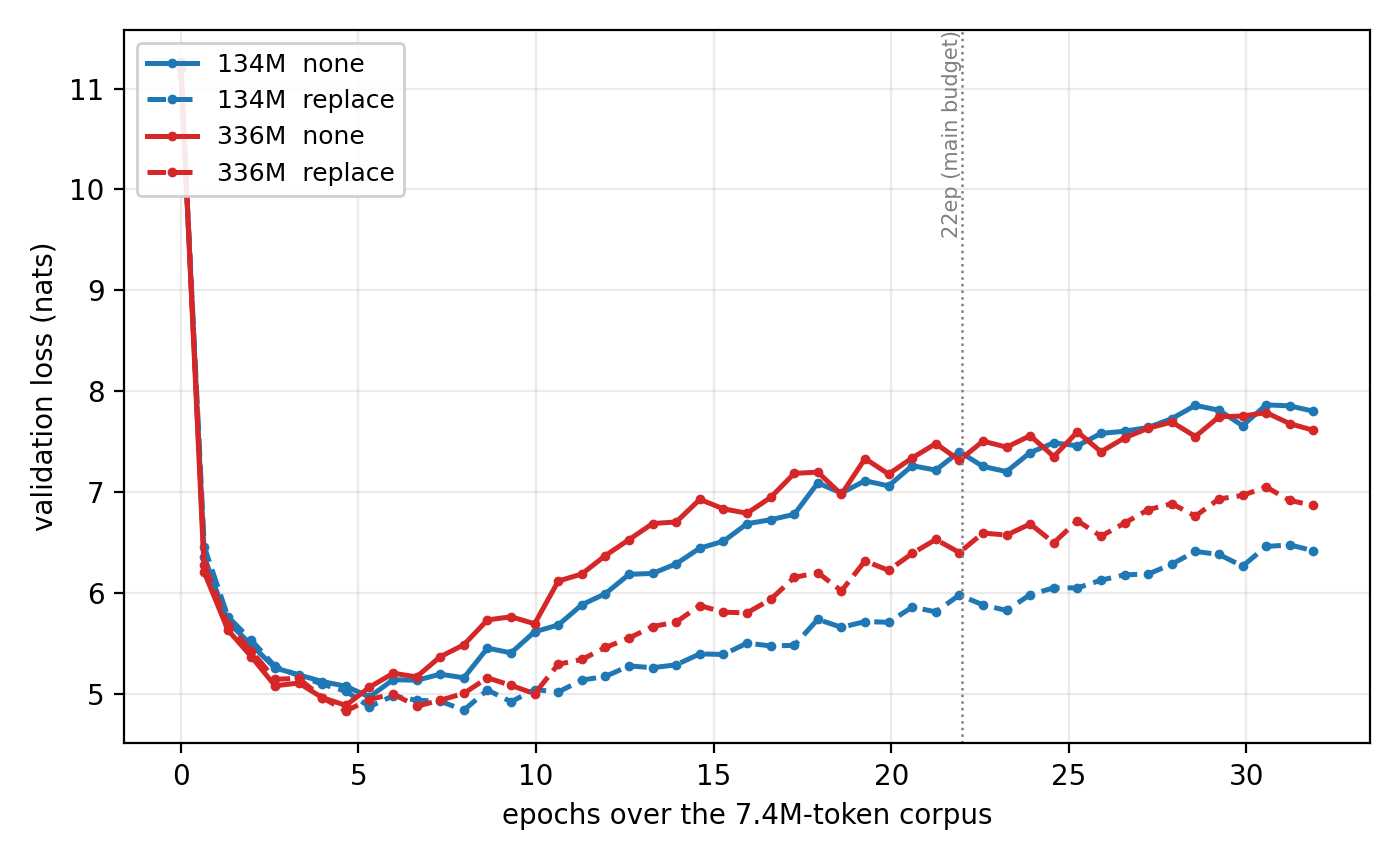}
\caption{Validation loss in nats over epochs on Korean web, one seed. Solid lines are \textsc{none} and dashed \textsc{replace}; blue is 134M and red 336M. The dotted line marks the $22$-epoch main budget.}
\label{fig:wall}
\end{figure}

To confirm that our corpora sit in the repeated-data regime, we extend one corpus (Korean web, $7.4$M tokens) to $32$ epochs under the fixed recipe and read validation loss as a function of epoch (Figure~\ref{fig:wall}). On the baseline (Table~\ref{app:tab:wall}) the validation loss reaches its minimum within about five epochs and then rises steeply, a $2.7$--$2.8$ nat increase by $32$ epochs. It places the $22$-epoch main budget well past the minimum. Replacement reaches an equal or slightly lower minimum and reduces the subsequent rise at both scales, from $2.73$ to $2.04$ nats at 336M and from $2.83$ to $1.58$ at 134M, the overfitting suppression of \S\ref{sec:safe} seen along the epoch axis.

\begin{table}[h]
\centering\small
\begin{tabular}{llccc}
\toprule
Scale & Cond & $L_{\text{min}}$ & epoch at min & final ($32$ep) \\
\midrule
134M & none & $4.97$ & $5.3$ & $7.80$ \\
134M & replace & $4.84$ & $8.0$ & $6.42$ \\
336M & none & $4.89$ & $4.7$ & $7.61$ \\
336M & replace & $4.83$ & $4.7$ & $6.87$ \\
\bottomrule
\end{tabular}
\caption{Epoch sweep on Korean web, $32$ epochs at the fixed recipe, one seed; \emph{final} is the loss at $32$ epochs. With one seed, the epoch of the minimum can differ from the three-seed means of Table~\ref{tab:seed}.}
\label{app:tab:wall}
\end{table}

\end{document}